\documentclass[10pt, a4paper]{article}

\usepackage[final]{lrec2026}

\usepackage{booktabs}
\usepackage{graphicx}
\usepackage{multirow}
\usepackage{xcolor}
\usepackage{adjustbox}
\usepackage{placeins}
\usepackage{float}

\title{Distilling Human-Aligned Privacy Sensitivity Assessment\\ from Large Language Models}

\name{
    Gabriel Loiseau$^{1,2}$, 
    {\bf \large Damien Sileo$^{2}$,}
    {\bf \large Damien Riquet$^{1}$,}\\
    {\bf \large Maxime Meyer$^{1}$,}
    {\bf \large Marc Tommasi$^{2}$}
}

\address{$^{1}$Hornetsecurity, Hem, France \\ $^2$Univ. Lille, Inria, CNRS, Centrale Lille, UMR 9189 - CRIStAL, F-59000 Lille, France \\
gabriel.loiseau@inria.fr}

\abstract{
Accurate privacy evaluation of textual data remains a critical challenge in privacy-preserving natural language processing. Recent work has shown that large language models (LLMs) can serve as reliable privacy evaluators, achieving strong agreement with human judgments; however, their computational cost and impracticality for processing sensitive data at scale limit real-world deployment. We address this gap by distilling the privacy assessment capabilities of Mistral Large 3 (675B) into lightweight encoder models with as few as 150M parameters. Leveraging a large-scale dataset of privacy-annotated texts spanning 10 diverse domains, we train efficient classifiers that preserve strong agreement with human annotations while dramatically reducing computational requirements. We validate our approach on human-annotated test data and demonstrate its practical utility as an evaluation metric for de-identification systems.
\\ \newline \Keywords{privacy evaluation, knowledge distillation, de-identification}}

\begin{document}

\maketitleabstract

\section{Introduction}
Quantifying privacy in textual data remains an open challenge due to the absence of a unified definition and the inherently contextual nature of privacy \citep{bambauer2022privacy, tesfay2016challenges}. Formal frameworks such as differential privacy \citep{dwork2006DP} provide rigorous guarantees, and proxy-based evaluation through attack success rates or information-theoretic measures are well-established in practice \cite{ren-etal-2025-measure}. However, these approaches capture specific, well-defined threat models rather than the broader, human-perceived notion of what constitutes sensitive content. Large language models (LLMs), with their capacity for nuanced language understanding, have emerged as promising candidates for human-aligned evaluation, demonstrating strong agreement with human judgments across a variety of NLP tasks \citep{zheng2023judging, li2024llmsasjudges}.


\begin{table*}[t]
\centering
\begin{adjustbox}{width=\textwidth}
\begin{tabular}{clp{11cm}r}
\hline
\textbf{Rating} & \textbf{Name} & \textbf{Description} & \textbf{Dataset (\%)} \\
\hline
1 & Harmless & Completely free of any private or sensitive information, including direct or indirect identifiers. & 46.01 \\
2 & Mostly not private & May contain some indirect identifiers, but is largely free of sensitive or personal information. & 16.58 \\
3 & Somewhat private & Contains some direct or indirect identifiers and can be considered moderately personal. & 16.81 \\
4 & Very private & Contains several direct or indirect identifiers and clearly includes personal information. & 14.21 \\
5 & Extremely private & Contains highly sensitive personal information or direct identifiers. & 6.38 \\
\hline
\end{tabular}
\end{adjustbox}
\caption{Privacy rating annotation scheme and resulting dataset distribution}
\label{tab:privacy-annotation-stats}
\end{table*}

Recent work by \citet{meisenbacher2025llm} represents a significant step toward closing this gap in privacy evaluation by applying the \emph{LLM-as-a-Judge} paradigm to this domain. Across 10 datasets and 677 human annotators, they show that LLMs can approximate a “global human privacy perspective” with strong agreement to aggregated human ratings, even exceeding inter-human agreement. These findings suggest that LLMs can serve as practical, human-aligned privacy evaluators.

Yet, deploying frontier LLMs for privacy assessment poses two central challenges. First, their computational and financial costs limit large-scale use. Second, evaluating sensitive text through third-party APIs introduces additional privacy concerns, as the very data being assessed may not be shareable. This creates a paradox: using powerful external LLMs to evaluate privacy may itself compromise privacy constraints.

In this work, we address this deployment gap through \emph{knowledge distillation}. Using Mistral Large 3 \citep{mistral2025large} as a teacher model, we annotate 200{,}000 user-written texts with privacy sensitivity scores following the structured Likert-scale methodology of \citet{meisenbacher2025llm}. We then distill these judgments into lightweight encoder-based classifiers, enabling fast, local, and privacy-preserving inference. Our central research question is whether the privacy reasoning capabilities of LLMs can be transferred to smaller models without sacrificing alignment with human judgments.

We validate the distilled models on human-annotated test data and show that they can match the agreement of their teacher model with aggregated human ratings. Beyond benchmark validation, we demonstrate that distilled privacy evaluators can serve as scalable automatic metrics for quantifying privacy reduction in text de-identification systems\footnote{Models, code, and data are available at \url{https://github.com/gabrielloiseau/privacy-distillation}}. Our contributions are threefold:
\begin{enumerate}
    \item We curate a large corpus of 200{,}000 texts, automatically annotated for privacy sensitivity using a state-of-the-art open LLM.
    \item We distill these LLM-generated privacy judgments into lightweight encoder models that achieve strong agreement with human annotations ($\alpha = 0.737$), surpassing the teacher model’s own human alignment ($\alpha = 0.716$), while enabling efficient and fully local inference.
    \item We demonstrate that distilled privacy evaluators function as scalable automatic metrics for assessing privacy reduction in text de-identification systems, and outline how compact privacy models open new research directions for privacy-aware NLP evaluation and system design.
\end{enumerate}

\section{Related Work}

\paragraph{Privacy Evaluation in NLP.}
In privacy-preserving NLP, evaluation commonly relies on proxy metrics such as re-identification success rates, simulated attacks, plausible deniability, or semantic similarity measures \citep{SHAHRIAR2025104358}. While these metrics capture specific threat models, they do not directly reflect how humans perceive the sensitivity of a text. Complementary research on automated privacy policy analysis \citep{wilson2016creation} and anonymization benchmarks \citep{lison-etal-2021-anonymisation, pilan-etal-2022-text, loiseau-etal-2025-tau} provides structured evaluation frameworks, primarily focusing on entity-level redaction quality. However, these approaches do not measure text-level privacy sensitivity or its alignment with human judgment across domains. Recent work has proposed \emph{LLM-as-a-Judge} as a scalable alternative to human evaluation for many NLP tasks \citep{li2024llmsasjudges, chiang-lee-2023-large, bavaresco-etal-2025-llms, li-etal-2025-generation}, with potential for modeling perceived privacy risk \citep{meisenbacher2025llm}.


\paragraph{LLM Distillation.}
Knowledge distillation \citep{hinton2015distilling} transfers capabilities from large teacher models to smaller student models. In NLP, it has produced efficient transformer variants such as DistilBERT \citep{sanh2019distilbert} and compressed generative LLMs into lightweight classifiers. Distillation can also rely solely on predicted labels, enabling black-box knowledge transfer when logits are unavailable \citep{chen2024knowledgedistillationblackboxlarge}. 

\section{Methodology}
\subsection{Privacy Annotation Framework}
We adopt the five-point Likert privacy sensitivity scale introduced by \citet{meisenbacher2025llm} and detailed in Table~\ref{tab:privacy-annotation-stats}, ranging from 1 (\textit{Harmless}) to 5 (\textit{Extremely private}). The scale operationalizes text-level privacy sensitivity by considering both direct identifiers (e.g., names, contact details) and broader contextual signals, including topical sensitivity (e.g., health conditions, legal situations), self-disclosure of personal experiences, and indirect identifiers that could enable re-identification in context. Sensitivity under this scale is therefore not limited to the presence of named entities or demographic attributes, but also encompasses the overall nature and intimacy of the disclosed content. The scale was previously validated through a large-scale human annotation study, and the resulting survey data were publicly released, providing a human-aligned target for supervision.

\subsection{LLM Annotation and Distillation Pipeline}
\paragraph{Data.}
We construct a corpus from the 10 publicly available datasets of user-written text from the original study spanning diverse domains: Blog Authorship Corpus (BAC), Enron Emails (EE), Medical Questions (MQ), Reddit Confessions (RC), Reddit Legal Advice (RLA), Mental Health Blog (MHB), Reddit Mental Health Posts (RMHP), Trustpilot Reviews (TR), Twitter (TW), and Yelp Reviews (YR). We sample 20{,}000 texts per dataset, excluding those used in the original human benchmark, resulting in approximately 200{,}000 texts. All data is in English; extending the approach to other languages remains future work. Additional details about each dataset are reported in Appendix~\ref{apx:dataset_sources}. 

\begin{table}[t]
\centering
\small
\begin{tabular}{lrrr}
\toprule
\textbf{Dataset} & \textbf{Avg tokens} & \textbf{$\bar{S}$} & \textbf{\% Priv.} \\
\midrule
MHB  &  272 & 3.31 & 77.4 \\
RMHP &  225 & 2.92 & 64.7 \\
RC   &  306 & 2.89 & 60.6 \\
RLA  &  283 & 2.84 & 60.3 \\
MQ   &  136 & 2.34 & 44.6 \\
EE   &  332 & 2.16 & 33.5 \\
BAC  &  264 & 1.78 & 23.5 \\
TR   &  69 & 1.28 &  5.2 \\
YR   &  129 & 1.21 &  2.6 \\
TW   &  50 & 1.11 &  1.7 \\
\midrule
\textbf{All}  & \textbf{207} & \textbf{2.18} & \textbf{37.4} \\
\bottomrule
\end{tabular}
\caption{Per-dataset statistics, sorted by mean privacy score. Avg tokens: mean BPE token count. $\bar{S}$: mean teacher-assigned privacy score (1--5). \% Priv.: percentage of texts rated $\geq 3$ (Somewhat private or above).}
\label{tab:dataset-breakdown}
\end{table}

\begin{table*}[t]
\centering
\small
\begin{adjustbox}{width=\textwidth}
\begin{tabular}{p{1.1cm}cp{16.0cm}}
\toprule
\textbf{Domain} & \textbf{Rating} & \textbf{Example} \\
\midrule
TW & 1 & ``Happy First Day of Spring!'' \\
TR & 2 & ``I have not received my item, even though I had an email from Royal Mail stating it had been delivered on 27/12/24'' \\
EE & 3 & ``Rick, Attached are drafts of the letters notifiying Williams, Apache and Lone Star that ENA will be acting as agent for Tenaska IV. Please review the letters and get back to me'' \\
MQ & 4 & ``Hi, Im asking for a friend that went to the hospital today by ambulance. she is 61 years old and has been sick for over a month.'' \\
MHB & 5 & ``I have suffered severe anxiety and depression for a very long time (first panic attack eight years old). I've seen plenty of psychologists ect but nothing seems to work.'' \\
\bottomrule
\end{tabular}
\end{adjustbox}
\caption{Examples illustrating the range of privacy ratings across domains.}
\label{tab:examples}
\end{table*}

\paragraph{Teacher Model Annotation.}
We use the open-weight Mistral Large 3 \citep{mistral2025large} as a teacher model to assign privacy sensitivity scores. We employ the structured prompting strategy of \citet{meisenbacher2025llm}, which provides explicit scale definitions and enforces discrete ratings. The full annotation prompt is provided in the Appendix~\ref{apx:prompt}. This yields a large, automatically labeled dataset reflecting LLM-based privacy judgments.

Table~\ref{tab:privacy-annotation-stats} shows the target rating distribution of the resulting dataset. The class distribution is notably imbalanced: nearly half of the texts (46\%) are rated as harmless, while the most sensitive category accounts for only about 6\% of samples, reflecting the natural scarcity of highly private content in everyday online communication. 

Table~\ref{tab:dataset-breakdown} provides a per-dataset breakdown, revealing variation in both text length and privacy sensitivity across domains. Health and confession-oriented domains (MHB, RC, RMHP, RLA) contain the highest proportions of private content, driven by self-disclosure of personal experiences, medical conditions, and sensitive life events. In contrast, review and microblog platforms (TR, TW, YR) are overwhelmingly rated as harmless (less than 6\% rated somewhat private or above), consistent with their public-facing, non-personal communication norms. Intermediate domains such as emails (EE) and blog posts (BAC) reflect a mixture, where privacy signals arise from incidental identifiers (names, contact details) rather than topical sensitivity. This diversity is essential for training a privacy evaluator that generalizes across the contextual factors that shape perceived sensitivity. Table~\ref{tab:examples} provides examples illustrating each rating level across different domains.

\paragraph{Student Models.}
We distill these annotations into encoder-based classifiers trained for 5-class classification. We evaluate four models: Ettin-150M, Ettin-17M \citep{weller2025ettin}, BERT-base \citep{devlin-etal-2019-bert}, and ModernBERT-base \citep{warner2024modernbert}. 
All models are fine-tuned using the same training recipe: learning rate $2 \times 10^{-5}$ with 10\% linear warmup, batch size 16, and 3 epochs. Due to its large size, the dataset is split into 90\% training, 5\% validation, and 5\% test sets. We select the best checkpoint by validation macro F1.

\section{Experiments}

We evaluate whether distilled encoder models can (1) learn the LLM-defined privacy task, and (2) preserve alignment with human privacy judgments. Our evaluation therefore combines standard classification metrics on our held-out test set with agreement-based analysis on the publicly released human benchmark from \citet{meisenbacher2025llm}. We quantify agreement using Krippendorff’s $\alpha$ \citep{krippendorff2011computing}, an inter-rater reliability coefficient defined as $\alpha = 1 - \frac{D_o}{D_e}$, where $D_o$ denotes the observed disagreement and $D_e$ the disagreement expected by chance; $\alpha=1$ indicates perfect agreement and $\alpha=0$ corresponds to chance-level agreement. We report two complementary agreement scores: agreement with the average human rating per text, and the average pairwise agreement with all individual annotations, for which we also report the standard deviation across annotators.

\begin{table*}[t]
\centering
\small
\begin{tabular}{llccc ccccc}
\toprule
\textbf{Model} & \textbf{Params} & \textbf{Acc. $\uparrow$} & \textbf{Macro F1 $\uparrow$} & \textbf{MAE $\downarrow$} 
& \multicolumn{5}{c}{\textbf{Per-class F1 $\uparrow$}} \\
\cmidrule(lr){6-10}
& & & & & \textbf{C1} & \textbf{C2} & \textbf{C3} & \textbf{C4} & \textbf{C5} \\
\midrule
Ettin-150M & 149M & 74.9 & 68.1 & 0.28 & 91.5 & 58.3 & 58.5 & 63.6 & 68.6 \\
ModernBERT-base & 149M & 73.7 & 67.2 & 0.28 & 91.9 & 57.4 & 57.7 & 62.8 & 68.2 \\
BERT-base & 110M & 73.3 & 65.9 & 0.29 & 91.3 & 55.4 & 55.7 & 60.6 & 66.4 \\
Ettin-17M & 17M & 71.1 & 62.5 & 0.34 & 90.1 & 51.2 & 52.8 & 57.5 & 60.7 \\
\bottomrule
\end{tabular}
\caption{Classification performance on the held-out test set. Accuracy and F1 scores are in \%. MAE is the mean absolute error on the 1--5 ordinal scale (range: 0--4). All encoder models are trained with identical hyperparameters. For reference, a Majority Class baseline achieves 45.9\% accuracy (12.5 macro F1), while a Random baseline achieves 20.0\% accuracy (18.0 macro F1).}
\label{tab:classification}
\end{table*}

\subsection{Learning the Distilled Task}
We first assess how well models learn the 5-class privacy classification task on our held-out test set. Table~\ref{tab:classification} reports accuracy, macro F1, mean absolute error\footnote{Mean Absolute Error (MAE) captures the average absolute distance between predicted and true privacy levels, treating the task as ordinal. Unlike accuracy or F1, MAE penalizes predictions proportionally to how far they deviate from the correct class (e.g., predicting 5 instead of 1 is penalized more than predicting 2 instead of 1), making it particularly suitable for ordered label spaces such as Likert-scale privacy ratings.}, and per-class F1 scores.

The Ettin-150M model achieves 74.9\% accuracy and 68.1 macro F1, substantially outperforming majority (45.9\%) and random (20.0\%) baselines. Performance is strong across the full privacy spectrum, including the most sensitive class (C5), where F1 reaches 68.6. Results are comparable to ModernBERT-base, while clearly surpassing BERT-base and the smaller Ettin-17M variant. 
F1 for the intermediate classes (C2--C4, ranging from 58 to 64) is lower than for the extreme classes (C1 at 91.5, C5 at 68.6). This is expected for an ordinal scale where adjacent categories are difficult to distinguish: texts at the boundary of ``Mostly not private'' and ``Somewhat private'' are inherently ambiguous. Importantly, classification errors for these middle classes are predominantly adjacent-class confusions (e.g., predicting C2 instead of C3), as reflected in the low MAE.
Overall, these results confirm that privacy sensitivity, as defined by the teacher model, can be learned reliably by lightweight encoders without architecture-specific tuning.

\subsection{Alignment with Human Privacy Judgments}
\begin{table}[t]
\centering
\small
\begin{adjustbox}{width=\columnwidth}
\begin{tabular}{lc}
\toprule
\textbf{Comparison} & \textbf{Krippendorff's $\alpha$} \\
\midrule
Ettin-150M vs.\ Human avg & 0.737 \\
Ettin-150M pairwise w/ humans & 0.514 ($\pm$0.265) \\
Ettin-17M vs.\ Human avg & 0.708 \\
Ettin-17M pairwise w/ humans & 0.498 ($\pm$0.259) \\
\midrule
Mistral Large 3 (675b) vs.\ Human avg & 0.716 \\
Mistral Large 3 (675b) pairwise w/ humans & 0.502 ($\pm$0.264) \\
Mistral-7b vs.\ Human avg & 0.563 \\
Mistral-7b pairwise w/ humans & 0.409 ($\pm$0.250) \\
\midrule
Inter-Human (overall) & 0.39 \\
Inter-Human (pairwise avg) & 0.54 \\
\bottomrule
\end{tabular}
\end{adjustbox}
\caption{Agreement metrics on the 250-text benchmark from \citet{meisenbacher2025llm}. ``Human avg'' denote agreement with the average rating of all human annotators.}
\label{tab:agreement}
\end{table}

We next evaluate agreement with the 677 human annotations from the original benchmark, which covers 250 texts in total (25 from each dataset). Table~\ref{tab:agreement} presents the central findings. The distilled Ettin-150M model achieves $\alpha = 0.737$ agreement with the \emph{average human rating}. Notably, this exceeds the agreement of its teacher model, Mistral Large 3 ($\alpha = 0.716$). For completeness, we also report results for Mistral-7b \cite{jiang2023mistral7b}, which achieves substantially lower agreement compared to both Mistral Large 3 and our distilled encoder models.

When compared pairwise with individual human annotators, the model achieves $\alpha = 0.514$ ($\pm0.265$), closely matching the inter-human pairwise average ($\alpha = 0.54$). This suggests that disagreements between the model and individual humans are of the same magnitude as disagreements among humans themselves. Our models align well on the general perception of privacy, whereas they cannot capture the unique perspectives and experiences of all represented annotators.

\subsection{De-Identification Evaluation}
To demonstrate a practical application, we evaluate our model's ability to assess anonymization quality using the Text Anonymization Benchmark (TAB) \citep{pilan-etal-2022-text}. TAB comprises English-language court cases from the European Court of Human Rights, with expert annotations of entity mentions categorized as \textsc{direct} identifiers (e.g., names, passport numbers), \textsc{quasi}-identifiers (e.g., age, nationality, profession), or \textsc{no\_mask}. Using the 555-document test split, we create four versions of each document: original, \textsc{direct}-masked (1{,}612 entities replaced with \texttt{[REDACTED]}), \textsc{quasi}-masked (19{,}197 entities), and fully masked (both types).

Table~\ref{tab:pseudonymization} reveals three key patterns. First, masking \textsc{direct} identifiers ($\Delta = 0.34$) has a larger per-entity effect than masking \textsc{quasi}-identifiers ($\Delta = 0.23$), despite far fewer entities (1{,}612 vs.\ 19{,}197), yielding higher privacy impact per entity for \textsc{direct} identifiers. This aligns with established personally identifiable information (PII) categorizations: names and other direct identifiers are inherently more individualizing than demographic attributes.

Second, comprehensive masking ($\Delta = 1.86$) produces a larger reduction than the sum of individual effects ($0.34 + 0.23 = 0.57$), revealing a strong interaction between identifier types. When both direct and quasi-identifiers are present, direct identifiers enable identification of the person's identity while quasi-identifiers provide additional sensitive information, thereby increasing the overall privacy risk of the text.

Third, after full masking, 84.1\% of documents are rated ``Harmless'' (class 1), compared to only 25.2\% in the original. This demonstrates that TAB's expert-defined masking scheme effectively reduces model-perceived privacy sensitivity. These results validate that our classifier captures privacy-relevant information consistent with expert annotations.

As a sanity check, we also randomly replace 30\% of words with \texttt{[REDACTED]} tokens. Rather than reducing privacy, random masking \emph{increases} the mean privacy score ($\bar{S} = 3.56$, $\Delta = -0.31$). This occurs because uninformed redaction disrupts coherence while preserving identifying content. This confirms that the classifier is sensitive to \emph{what} is masked, not merely to the presence of masking tokens.
\begin{table}[t]
\centering
\small
\begin{tabular}{lccc}
\toprule
\textbf{Condition} & \textbf{$\bar{S}$} & \textbf{$\Delta$} & \textbf{\% Class 1} \\
\midrule
Original            & 3.25 & -- & 25.2 \\
Mask \textsc{direct}  & 2.91 & 0.34 & 30.5 \\
Mask \textsc{quasi}   & 3.02 & 0.23 & 28.5 \\
Mask ALL            & 1.39 & 1.86 & 84.1 \\
\midrule
Mask 30\% random    & 3.56 & -0.31 & 17.3 \\
\bottomrule
\end{tabular}
\caption{Privacy scores on TAB test set. $\bar{S}$: mean score (1--5). $\Delta$: reduction from original. \% Class 1: proportion rated ``Harmless.''}
\label{tab:pseudonymization}
\end{table}

\section{Discussion \& Future Work}

\paragraph{Performance Measurement.}
A notable outcome is that the distilled Ettin-150M slightly exceeds the teacher model in agreement with the \emph{average} human rating. This does \emph{not} imply that the student is intrinsically ``more correct'' than the teacher using our approach; rather, distillation can act as a \emph{denoising} process. Training on a large volume of teacher-labeled examples can smooth prompt-level stochasticity and compress the teacher's reasoning into a deterministic decision boundary that generalizes better on a small human benchmark. Future work should explicitly test this hypothesis by studying more teacher behaviors or varying the amount of distillation data.

\paragraph{Use cases.}
Beyond benchmarking, an on-device privacy sensitivity classifier unlocks workflows that are difficult or undesirable with API-based LLM judges:
(i) \textit{Dataset curation}: assigning sensitivity scores to large corpora to route high-risk examples for manual review, filtering, or access control before model training;
(ii) \textit{Privacy-aware evaluation for rewriting/anonymization}: using the score as an automatic metric to compare de-identification or privatization systems across datasets and parameter settings, complementing attack-based proxies;
(iii) \textit{User-facing privacy assistance}: real-time warnings in writing assistants (e.g., ``this message contains likely identifying details'') and suggestions for minimal edits. This is particularly valuable given evidence that users routinely leak PII when interacting with external LLMs~\cite{mireshghallah2024trust}.

\paragraph{Future Work.}
Compact evaluators enable research that is otherwise compute- or policy-constrained. First, they make it feasible to study \textit{privacy signals at scale} via attribution and counterfactual edits, helping disentangle identifiability cues (names, locations, unique events) from topic sensitivity (health, legal issues, mental health). Second, the model can serve as a \textit{training signal}: combined with a utility measure (e.g., semantic similarity), it can define privacy--utility trade-offs and support search or learning procedures that find minimal changes that reduce privacy sensitivity. Third, future work should move beyond a single ``global'' notion of privacy by incorporating \textit{context} (audience, purpose, setting) and exploring \textit{personalization} with small amounts of user-provided preference data. Finally, robustness remains open: calibrating scores, dealing with out-of-domain inputs, and auditing domain- and demographic-dependent failure modes are essential before deploying the model as part of automated pipelines.

\section{Conclusion}
We presented a knowledge distillation approach for creating efficient privacy sensitivity classifiers from LLM judgments. Responding to calls for lightweight privacy evaluation models, we distilled Mistral Large's privacy assessments into a 150M-parameter Ettin encoder that achieves strong agreement with human annotations while enabling private and faster inference. Our evaluation on the Text Anonymization Benchmark demonstrates that the classifier captures meaningful differences between direct and quasi-identifiers in expert-annotated documents, validating its utility for de-identification assessment. We release our code, models and dataset to support reproducible privacy evaluation in NLP.

\section*{Limitations}
Our models inherit the privacy notion and potential biases of the teacher LLM: privacy is compressed into a single 1--5 sensitivity score, which may conflate multiple dimensions such as identifiability and topic sensitivity. The training data is English-only; multilingual transfer remains untested. Privacy is contextual \citep{nissenbaum2004privacy}, yet the classifier evaluates texts largely in isolation without explicit information about audience, purpose, or setting.

Teacher labeling can be stochastic due to the non-deterministic nature of large language models \citep{song-etal-2025-good}; using multiple teachers, or a small amount of human-labeled calibration data could reduce noise and improve robustness. We also did not systematically study alternative teacher models or distillation strategies. Finally, the score should not be interpreted as a formal privacy guarantee or as a proxy for adversarial re-identification risk: it captures \emph{perceived} sensitivity under the adopted scale.

\section*{Ethical Considerations}
This work processes potentially sensitive user-generated content. 
All source datasets are publicly available and have been previously used in research. 
Our classifier is intended for evaluating privacy-preserving methods and supporting privacy research, not for making decisions about individuals or for surveillance purposes. 
We caution against using the model as a hard gate without human oversight: the privacy scale is a subjective construct, and model scores should inform rather than replace human judgment.

\section*{Acknowledgments}
We thank the anonymous reviewers for their constructive feedback. We are also grateful to the authors of \citet{meisenbacher2025llm}, whose work on LLM-based privacy evaluation provided the foundation for this study.

\section*{Bibliographical References}\label{sec:reference}

\bibliographystyle{lrec2026-natbib}
\bibliography{references,anthology-1,anthology-2}

\newpage
\appendix
\section{Dataset Sources Description}
\label{apx:dataset_sources}
To construct a diverse and coherent valid privacy corpus, we aggregate user-written texts from ten publicly available datasets spanning blogs, emails, health-related forums, legal advice, reviews, microblogs, and online discussions. Together, these datasets cover varying genres, platform norms, and disclosure styles, enabling us to capture multiple dimensions of privacy risk: explicit identifiers, contextual clues, health and legal sensitivity, and stylometric linkability.

\textbf{Blog Authorship Corpus (BAC).}
BAC consists of long-form blog posts, capturing personal storytelling and opinionated writing with strong stylistic signatures---a setting commonly associated with authorship and user profiling tasks. Such stylometric consistency makes it a natural domain for studying linkability-based privacy risks, beyond overt identifiers. We use the HuggingFace version of the corpus (\url{https://hf.co/datasets/tasksource/blog_authorship_corpus}).

\textbf{Enron Emails (EE).}
EE comprises semi-private organizational communication, where texts often include operational details, social relations, and direct contact information. This domain contributes a complementary privacy profile: not necessarily self-disclosure, but high likelihood of concrete identifiers and workplace context. We use the HuggingFace release (\url{https://hf.co/datasets/sujan-maharjan/enron_email_dataset}), totaling 362{,}180 emails from the sent items folder.

\textbf{Medical Questions (MQ).}
The Medical Question Answering dataset contains over one million user-written medical question-answer pairs; we keep only the questions. These texts are representative of information-seeking and triage-like tasks in health NLP, and they frequently contain sensitive attributes (symptoms, medications, ages, conditions), making MQ a core pillar of the corpus for health-related privacy. We use the processed HuggingFace version (\url{https://hf.co/datasets/Malikeh1375/medical-question-answering-datasets}), comprising 246{,}678 questions.

\textbf{Mental Health Blog (MHB).}
MHB contains posts from a mental health forum (2011--2020), reflecting long-form self-disclosure and peer support---a genre often used for mental-health-related classification and risk detection. Privacy signals here are typically driven by sensitive topics and personal experiences, even when explicit identifiers are absent \cite{boinepelli-etal-2022-leveraging}. (This dataset is not hosted on HuggingFace.)

\textbf{Reddit Confessions (RC).}
This corpus (\url{https://hf.co/datasets/SocialGrep/one-million-reddit-confessions}) consists of one million posts from four confession-oriented subreddits, emphasizing anonymous self-disclosure and narrative accounts. As a domain, it broadens our corpus toward ``voluntary disclosure'' settings, where privacy sensitivity is often rooted in intimate content rather than formal identifiers.

\textbf{Reddit Legal Advice (RLA).}
RLA contains nearly 100k informal legal help-seeking posts from \textit{/r/legaladvice} \cite{li-etal-2022-parameter}. We use the HuggingFace version (\url{https://hf.co/datasets/jonathanli/legal-advice-reddit}), which provides 79{,}136 posts. This domain introduces a distinct mix of sensitive situational details (disputes, employment issues, alleged wrongdoing) and contextual clues (locations, timelines, involved parties).

\textbf{Reddit Mental Health Posts (RMHP).}
RMHP provides over 150k posts from five mental-health-related subreddits. Compared to MHB, RMHP adds a different platform style (shorter, conversational, often more fragmented) while covering similar sensitive themes. This contrast helps ensure our privacy corpus captures both topic sensitivity \emph{and} platform-/genre-specific expressions of disclosure. We use the HuggingFace version (\url{https://hf.co/datasets/solomonk/reddit_mental_health_posts}), comprising 151{,}288 posts.

\textbf{Trustpilot Reviews (TR).}
Trustpilot reviews are consumer-facing evaluations of services and products, commonly used in sentiment and sociolinguistic analyses \cite{10.1145/2736277.2741141}. While reviews may appear ``less private'' on average, they regularly include incidental identifiers (names, locations, service interactions) and thus provide valuable coverage of \emph{everyday} privacy leakage in ostensibly public feedback. We use the HuggingFace dataset (\url{https://hf.co/datasets/Kerassy/trustpilot-reviews-123k}), containing 123{,}181 \textit{en-US} reviews.

\textbf{Twitter (TW).}
TW represents short, public microblog posts, often used for topic, sentiment, stance, or event-oriented analyses. Including Twitter ensures the corpus captures privacy signals in highly compressed text, where identifying cues can arise from usernames, mentions, location hints, or contextual references rather than explicit PII. We sample from the HuggingFace version of the 100M tweets corpus (\url{https://hf.co/datasets/enryu43/twitter100m_tweets}), containing 88{,}084{,}332 tweets.

\textbf{Yelp Reviews (YR).}
YR contains user reviews of local businesses---a domain frequently associated with sentiment and recommendation tasks, grounded in real-world places and experiences. This setting contributes privacy risks tied to location context, routines, and named venues (even without explicit identifiers). We use the HuggingFace release (\url{https://hf.co/datasets/yashraizad/yelp-open-dataset-reviews}), which contains 6{,}990{,}280 reviews.

\section{Teacher Prompt}
\label{apx:prompt}
We use a structured prompt that enforces the discrete 1--5 privacy scale:

\begin{quote}\small
Provide a \textbf{privacy rating} scoring on how sensitive or private the user\_text, for example including personal information or indirectly identifiable information.

Give your answer on a scale of 1 to 5, where 1 means that the user\_text is not sensitive or private at all, and 5 means that text contains highly sensitive or private information.

Here is the scale you should use to build your answer:\\
1: The user\_text is harmless: completely free of any private or sensitive information, either direct or indirect identifiers.\\
2: The user\_text is mostly not private: may contain some indirect identifiers, but is mostly free of sensitive information.\\
3: The user\_text is somewhat private: contains some direct or indirect identifiers, and can be deemed somewhat personal information.\\
4: The user\_text is very private: contains several direct or indirect identifiers, and it is clear that this text contains personal information.\\
5: The user\_text is extremely private: contains highly sensitive information, such as direct personal identifiers, and the text is clearly something that should not be made public.

\end{quote}

\end{document}